\documentclass[review]{elsarticle}
\usepackage{amsmath,amssymb,amsfonts,bm}
\usepackage{stfloats}
\usepackage{CJKutf8}
\usepackage[linesnumbered,lined,boxed,commentsnumbered]{algorithm2e}
\usepackage[noend]{algpseudocode}
\usepackage{multirow}
\usepackage{lineno}
\usepackage[noend]{algpseudocode}
\usepackage{graphicx}
\usepackage{array}
\usepackage{multirow}
\usepackage{amsfonts}
\usepackage[dvipsnames]{xcolor}
\usepackage{booktabs}
\usepackage{textcomp}
\usepackage{scrextend}
\usepackage{subfig}

\makeatletter
\providecommand{\doi}[1]{%
  \begingroup
    \let\bibinfo\@secondoftwo
    \urlstyle{rm}%
    \href{http://dx.doi.org/#1}{%
      doi:\discretionary{}{}{}%
      \nolinkurl{#1}%
    }%
  \endgroup
}
\makeatother

\makeatletter
\def\ps@pprintTitle{%
   \let\@oddhead\@empty
   \let\@evenhead\@empty
   \def\@oddfoot{\reset@font\hfil\thepage\hfil}
   \let\@evenfoot\@oddfoot
}
\makeatother

\begin{document}
\begin{CJK}{UTF8}{gkai}

\begin{frontmatter}

\title{Fine-tuning ERNIE for Chest Abnormal Imaging Signs Extraction}

\author{Zhaoning Li}
\ead{lizhn7@mail2.sysu.edu.cn}
\author{Jiangtao Ren\corref{mycorrespondingauthor}}
\ead{issrjt@mail.sysu.edu.cn}
\address{School of Data and Computer Science, Guangdong Province Key Lab of Computational Science, Sun Yat-Sen University, Guangzhou, Guangdong 510006, PR China}
\cortext[mycorrespondingauthor]{Corresponding author}

\begin{abstract}
Chest imaging reports describe the results of chest radiography procedures. Automatic extraction of abnormal imaging signs from chest imaging reports has a pivotal role in clinical research and a wide range of downstream medical tasks. However, there are few studies on information extraction from Chinese chest imaging reports. In this paper, we formulate chest abnormal imaging sign extraction as a sequence tagging and matching problem. On this basis, we propose a transferred abnormal imaging signs extractor with pretrained ERNIE as the backbone, named \textbf{EASON} (fine-tuning \textbf{\textcolor{red}{E}}RNIE with CRF for \textbf{\textcolor{red}{A}}bnormal \textbf{\textcolor{red}{S}}igns Extracti\textbf{\textcolor{red}{ON}}), which can address the problem of data insufficiency. In addition, to assign the attributes (the body part and degree) to corresponding abnormal imaging signs from the results of the sequence tagging model, we design a simple but effective \textbf{tag2relation} algorithm based on the nature of chest imaging report text. We evaluate our method on the corpus provided by a medical big data company, and the experimental results demonstrate that our method achieves significant and consistent improvement compared to other baselines.
\end{abstract}


\end{frontmatter}


\insert\footins{
 \normalfont\footnotesize
  \interlinepenalty\interfootnotelinepenalty
  \splittopskip\footnotesep \splitmaxdepth \dp\strutbox
  \floatingpenalty10000 \hsize\columnwidth
\begin{tabular}{|p{.9\textwidth}|}
 \hline
This is a post-print (accepted manuscript) version of this paper, which has been published in Journal of Biomedical Informatics. This work is licensed under a Creative Commons Attribution-NonCommercial-NoDerivatives 4.0 International License (CC-BY-NC-ND 4.0).
 \begin{center}
 \includegraphics[scale=0.35]{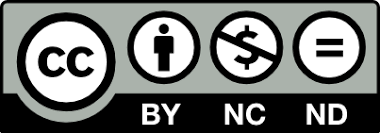}
 \end{center}\\
 \hline
 \end{tabular}}

\section{Introduction}

A large number of radiology reports have been accumulated for communication and documentation of diagnostic imaging since the wide use of medical information systems in China. In addition to the application of radiographs in medical image analysis \cite{DBLP:journals/mia/LitjensKBSCGLGS17,DBLP:journals/corr/abs-1811-10052}, radiology reports also contain considerable meaningful knowledge to be discovered, and harnessing their potential requires efficient and automated information extraction \cite{Pons_2016}. For example, the automatic extraction of abnormal imaging signs in chest imaging reports is essential for clinical research and a wide range of downstream medical tasks: patient similarity measuring \cite{DBLP:conf/cikm/Ni0ZYM17}, diagnosis prediction \cite{DBLP:conf/cikm/Ni0ZYM17} and automatic ICD coding \cite{DBLP:conf/naacl/MullenbachWDSE18}. However, most of the existing information extraction systems in radiology are developed for English \cite{Friedman_1995,Johnson_1997,DBLP:journals/jbi/EsuliMS13,DBLP:journals/jamia/BozkurtLSR15,DBLP:journals/artmed/HassanpourL16,DBLP:journals/jbi/GuptaBR18}, and too little work has been devoted to the extraction of abnormal imaging signs from Chinese chest imaging report text.

\begin{figure}[htbp]
\centerline{\includegraphics[scale=0.68]{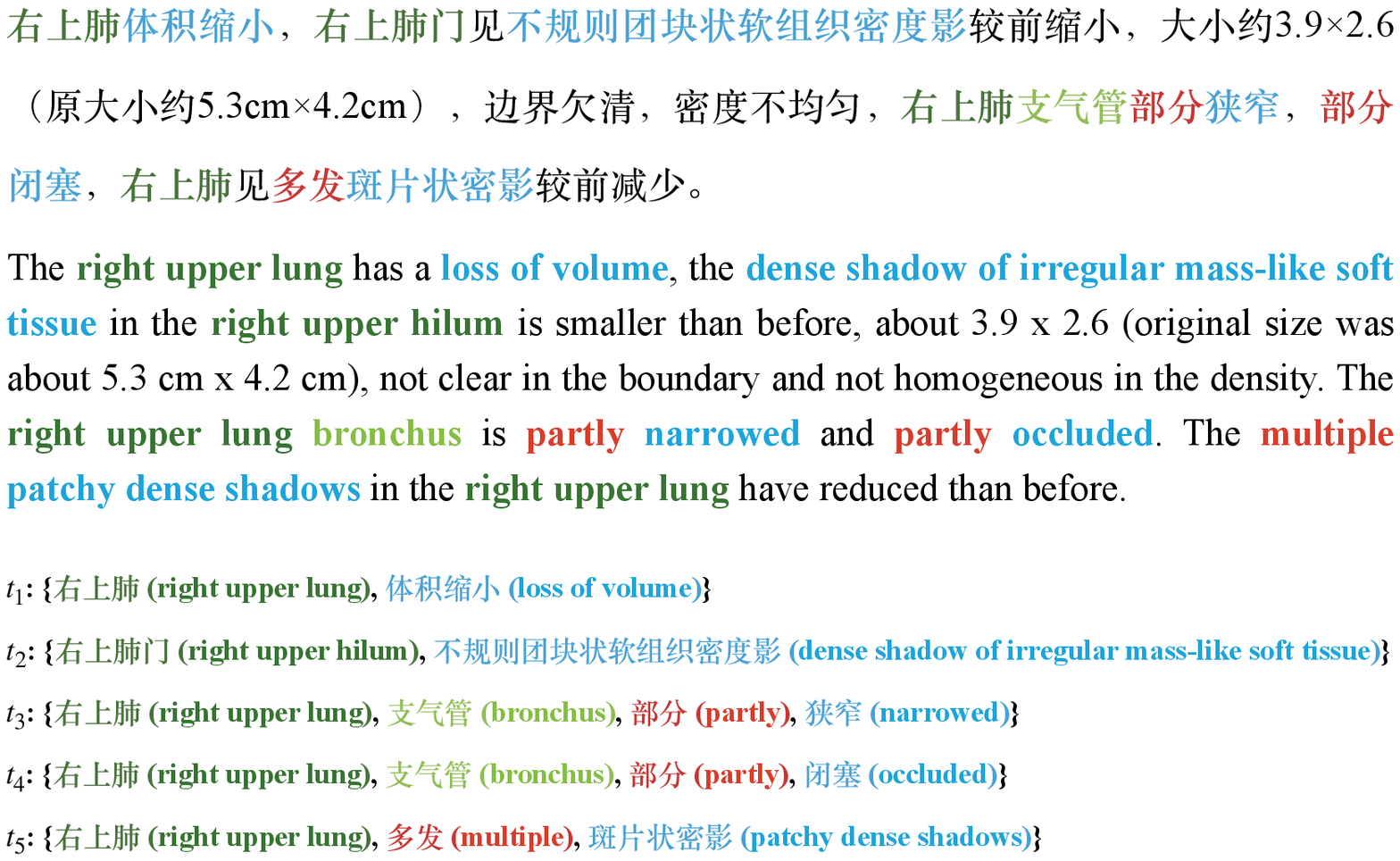}}
\caption{A standard example of chest abnormal imaging sign extraction. In this case, $t_{i}$ represents the i-th tuple in the above sentence; \textbf{\textcolor{Cyan}{cyan}} denotes the abnormal imaging signs, \textbf{\textcolor{Red}{red}} denotes the degree of the abnormal imaging sign, \textbf{\textcolor{OliveGreen}{olive green}} and \textbf{\textcolor{LimeGreen}{lime green}} denote the primary and secondary body part of the abnormal imaging sign, respectively.}
\label{fig1}
\end{figure}

In this paper, we aim to extract structured information of abnormal imaging signs (i.e., abnormal imaging signs and their attributes ``where did the abnormality occur'' and ``what is the degree of abnormality'') from unstructured chest imaging reports. For example, ``\textbf{\textcolor{Cyan}{闭塞}}'' (occluded) is an abnormal imaging sign; ``\textbf{\textcolor{OliveGreen}{右上肺}}'' (right upper lung) and ``\textbf{\textcolor{LimeGreen}{支气管}}'' (bronchus) are primary and secondary body parts where occlusion occurs, respectively; ``\textbf{\textcolor{Red}{部分}}'' (partly) is the degree to which occlusion occurs, as shown in Fig.~\ref{fig1}.

Specifically, we can divide this information extraction task into three subtasks:

\begin{enumerate}
\item Extracting abnormal imaging signs;
\item Extracting attributes of abnormal imaging signs;
\item Matching between abnormal imaging signs and their attributes.
\end{enumerate}

To accurately and efficiently extract abnormal imaging signs and their attributes from chest imaging reports, we formulate subtasks 1) and 2) into a sequence tagging problem at the Chinese character level, which can avoid introducing errors caused by segmentation. Traditionally, researchers use machine learning methods \cite{DBLP:conf/icml/McCallumFP00,DBLP:conf/acl/ZhouS02,DBLP:conf/conll/McCallum003} to perform sequence tagging tasks. Recently, with the development of deep learning, deep learning architectures based on long short-term memory networks (LSTMs) \cite{DBLP:journals/neco/HochreiterS97} or convolutional neural networks (CNNs) \cite{DBLP:journals/neco/LeCunBDHHHJ89} combined with conditional random fields (CRFs) \cite{DBLP:conf/icml/LaffertyMP01} have achieved state-of-the-art results in the clinical field \cite{DBLP:journals/bioinformatics/HabibiWNWL17,DBLP:journals/jbi/WangZRGXH19,DBLP:conf/bibm/QiuWZRG18}. In real medical situations, however, the cost of manually labeling a large training set in the medical field is too high and error-prone \cite{DBLP:conf/acl/ZhengWBHZX17,DBLP:journals/jbi/GuptaBR18}. In the case of the insufficiency of high-quality annotated medical data, the data-hungry nature of deep learning limits the performance of these neural-based models. In this work, we adopt the advanced deep sequence tagging framework to address the following questions:

\begin{itemize}
\item How can data insufficiency be alleviated?
\item How can attributes be assigned to abnormal imaging signs?
\end{itemize}

\textbf{First}, to alleviate the problem of data insufficiency, we propose 
fine-tuning ERNIE \cite{DBLP:journals/corr/abs-1904-09223} (Enhanced Representation through Knowledge Integration) (Section~\ref{sec3.3}) for our task, which is pretrained on a large corpus and has achieved ground-breaking performance across various Chinese natural language processing (NLP) tasks. Experimental results show that this transfer learning method can drastically improve the performance of our task.

\textbf{Second}, to assign the attributes to the corresponding abnormal imaging signs, i.e., subtask 3), we design a simple but effective \textbf{Tag2Relation} algorithm (Section~\ref{sec3.4}) based on the nature of chest imaging report text, which can easily construct the relation between entities from the results of the sequence tagging model.

The contributions of this paper can be summarized as follows:

\begin{enumerate}
\item We propose \textbf{EASON} (fine-tuning \textbf{\textcolor{red}{E}}RNIE with CRF for \textbf{\textcolor{red}{A}}bnormal \textbf{\textcolor{red}{S}}igns Extracti\textbf{\textcolor{red}{ON}}), a transferred chest abnormal imaging signs extractor. To the best of our knowledge, we are the first to present such an effective method for the automatic extraction of abnormal imaging signs from chest imaging reports.
\item We design a novel tag2relation algorithm to establish the relation between abnormal imaging signs and their attributes, which can easily match the abnormal imaging sign with their attributes based on the result of the sequence tagging model.
\item We conduct extensive experiments on chest imaging reports provided by a medical big data company. Experimental results (Section~\ref{sec4.5}) and further analysis (Section~\ref{sec5}) show that our method achieves significant and consistent improvement compared to other baselines. We release the code and terminology to the research community for further research \footnote{https://github.com/Das-Boot/eason}.
\end{enumerate}

\section{Related Work}

With the integration and development of medicine and computer science technology, clinical information extraction is becoming increasingly important and attracting increasing attention. Many clinical NLP systems have been developed to extract structured information from unstructured electronic health records (EHRs). The research methods for clinical information extraction mainly include rule-based methods, machine learning-based methods, and deep learning-based methods.

Rule-based methods are the earliest attempt to extract information from EHRs \cite{DBLP:journals/jamia/FriedmanAACJ94,Friedman_1995,Johnson_1997,DBLP:journals/midm/ZengGWSML06,DBLP:journals/jbi/CodenSSTMSCGG09,DBLP:journals/jbi/HarkemaDTC09,DBLP:journals/jamia/BozkurtLSR15}. For example, \citet{Friedman_1995} proposed MEDLEE (medical language extraction and encoding system) to extract information from textual patient reports with controlled vocabulary and grammatical rules. \citet{Johnson_1997} designed RADA (radiology analysis tool) to extract and structure key medical concepts and their attributes contained in radiology reports through predefined rules. \citet{DBLP:journals/jbi/HarkemaDTC09} proposed the ConText algorithm for determining whether clinical conditions mentioned in clinical reports are negated, hypothetical, historical, or experienced by someone other than the patient. ConText is based on the simple approach used by NegEx \cite{Chapman_2001} (a regular expression algorithm) for finding negated conditions in text. These rule-based methods require formulating rules that consume significant time and effort, and their limited coverage and generalizability are the main drawbacks.

The traditional machine learning-based methods include handcrafted features for hidden Markov models (HMMs) \cite{DBLP:conf/acl/ZhouS02,DBLP:journals/midm/SongYH15}, maximum entropy Markov models (MEMMs) \cite{DBLP:conf/icml/McCallumFP00,DBLP:conf/bionlp/FinkelDNNMS04}, CRFs \cite{DBLP:conf/conll/McCallum003,DBLP:journals/jbi/SkeppstedtKND14} and support vector machines (SVMs) \cite{DBLP:conf/pakdd/WuFLY06,ju2011named}. In other related work targeting radiology reports, \citet{DBLP:journals/jbi/EsuliMS13} performed information extraction from free-text radiology reports with the CRF-based method. \citet{DBLP:journals/artmed/HassanpourL16} used the conditional Markov model (CMM) and CRFs to extract radiological observations from reports. These machine learning-based methods heavily rely on a large number of feature engineering and thus require considerable human effort and time on feature engineering.

In recent years, deep learning has ushered in incredible advances in NLP tasks. Different from shallow machine learning methods, deep neural networks rely on powerful representation learning ability to automatically discover features, which significantly reduces feature engineering and saves human resources and time. For example, \citet{DBLP:journals/jbi/GuptaBR18} used an unsupervised model to extract relations and their associated entities from radiology reports using automated clustering of similar relations in narrative mammography radiology reports. The proposed approach based on distributional semantics (neural representation) and clustering to find similar relations outperforms other approaches. In addition, sequence tagging methods based on LSTMs or CNNs combined with a CRF \cite{DBLP:conf/icml/LaffertyMP01} layer achieve state-of-the-art performance in the clinical field and outperform traditional statistical methods \cite{DBLP:journals/bioinformatics/HabibiWNWL17,DBLP:journals/jbi/WangZRGXH19,DBLP:conf/bibm/QiuWZRG18}. However, the data-hungry nature of deep learning limits the performance of these neural-based methods for medical tasks with small datasets. The recent development of language representation models \cite{DBLP:conf/naacl/PetersNIGCLZ18,DBLP:conf/coling/AkbikBV18,DBLP:conf/naacl/DevlinCLT19,DBLP:journals/corr/abs-1904-09223,DBLP:journals/corr/abs-1906-08101,DBLP:journals/corr/abs-1907-12412} trained on a large corpus demonstrate the possibility of transfer learning for sequence tagging.

\section{Methods \label{sec3}}

\subsection{Problem Definition \label{sec3.1}}

To better illustrate our method, we introduce some terminologies as follows:

\begin{itemize}
\item \textbf{Abnormal imaging sign (\textit{Abn}):} An entity refers to abnormal results of chest radiographs, CT, MR, etc.
\item \textbf{Body part (\textit{P}):} An entity refers to the specific organ or tissue structure where the abnormal imaging sign occurs, which serves as an attribute of abnormal imaging signs, including the primary body part (\textbf{\textit{PP}}) and the secondary body part (\textbf{\textit{SP}}).
\item \textbf{Degree (\textit{D}):} An entity refers to the scope (e.g., ``弥漫'' (diffusely)), severity (e.g., ``轻度'' (slightly)), frequency (e.g., ``多发'' (multiple)), and quantity (e.g., ``单个'' (single)) of abnormal imaging sign occurrence, which also serves as an attribute of abnormal imaging signs \footnote{We only focus on the degree entities in the description of the current chest imaging report. We do not care about the degree entities related to historical condition, so we do not annotate ``较前减少'' (reduced than before) as degree in the sentence ``右上肺见多发斑片状密影较前减少。'' (The multiple patchy dense shadows in the right upper lung have reduced than before).}.
\end{itemize}

Then, the structured information of an abnormal imaging sign can be formally defined as a quadruple: \{\textit{PP}, \textit{SP}, \textit{D}, \textit{Abn}\}, where \textit{PP}, \textit{SP}, \textit{D} are attributes of the corresponding \textit{Abn}; note that the attribute may be null, as shown in Fig~\ref{fig1}.

Thus, in this work, our task is to extract all the quadruples in a given chest imaging report, which includes abnormal imaging sign \textit{Abn} identification, attribute \textit{PP}/\textit{SP}/\textit{D} identification, and matching between \textit{Abn} and \textit{PP}/\textit{SP}/\textit{D}.

\subsection{Tagging Scheme}

We use the ``BIO” (begin, inside, other) and ``Abn, P, D'' signs to represent the position information and the semantic roles of the Chinese characters, respectively. Note that we only label the \textit{P} or \textit{D} serving as an attribute of \textit{Abn}. Both the primary and secondary body parts are marked as different body parts; for example, ``右上肺支气管'' (right upper lung bronchi) is labeled as ``\textbf{\textcolor{OliveGreen}{右上肺}}'' (right upper lung) and ``\textbf{\textcolor{LimeGreen}{支气管}}'' (the bronchi), respectively.

\begin{figure}[htbp]
\centerline{\includegraphics[scale=0.145]{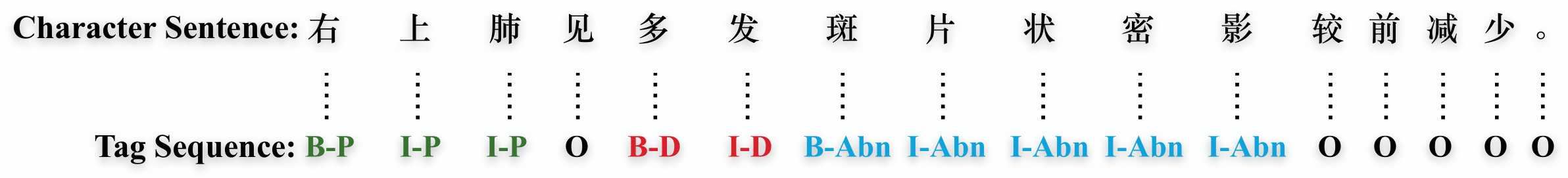}}
\caption{A standard annotation for the example sentence based on our tagging scheme.}
\label{fig2}
\end{figure}

Fig.~\ref{fig2} shows an example of such a tagging scheme for the sentence ``右上肺见多发斑片状密影较前减少。'' (The multiple patchy dense shadows in the right upper lung have reduced than before). Based on our tagging scheme, we can label the abnormal imaging sign: ``\textbf{\textcolor{Cyan}{斑片状密影}}'' (patchy dense shadows), body part: ``\textbf{\textcolor{OliveGreen}{右上肺}}'' (right upper lung) and degree: ``\textbf{\textcolor{Red}{多发}}'' (multiple) separately with our unique tags. Specifically, tag ``O'' represents the ``other'', which means that the corresponding character is irrelevant in any entity components. Tag ``B-P'' represents the ``body part begin'', tag ``I-P'' represents the ``body part inside'', tag ``B-D'' represents the ``degree begin'', tag ``I-D'' represents the ``degree inside'', tag ``B-Abn'' represents the ``abnormal imaging sign begin'' and tag ``I-Abn'' represents the ``abnormal imaging sign inside''.

\subsection{Extracting Abnormal Imaging Signs and Attributes with EASON \label{sec3.3}}

\begin{figure}[htbp]
\centerline{\includegraphics[scale=0.625]{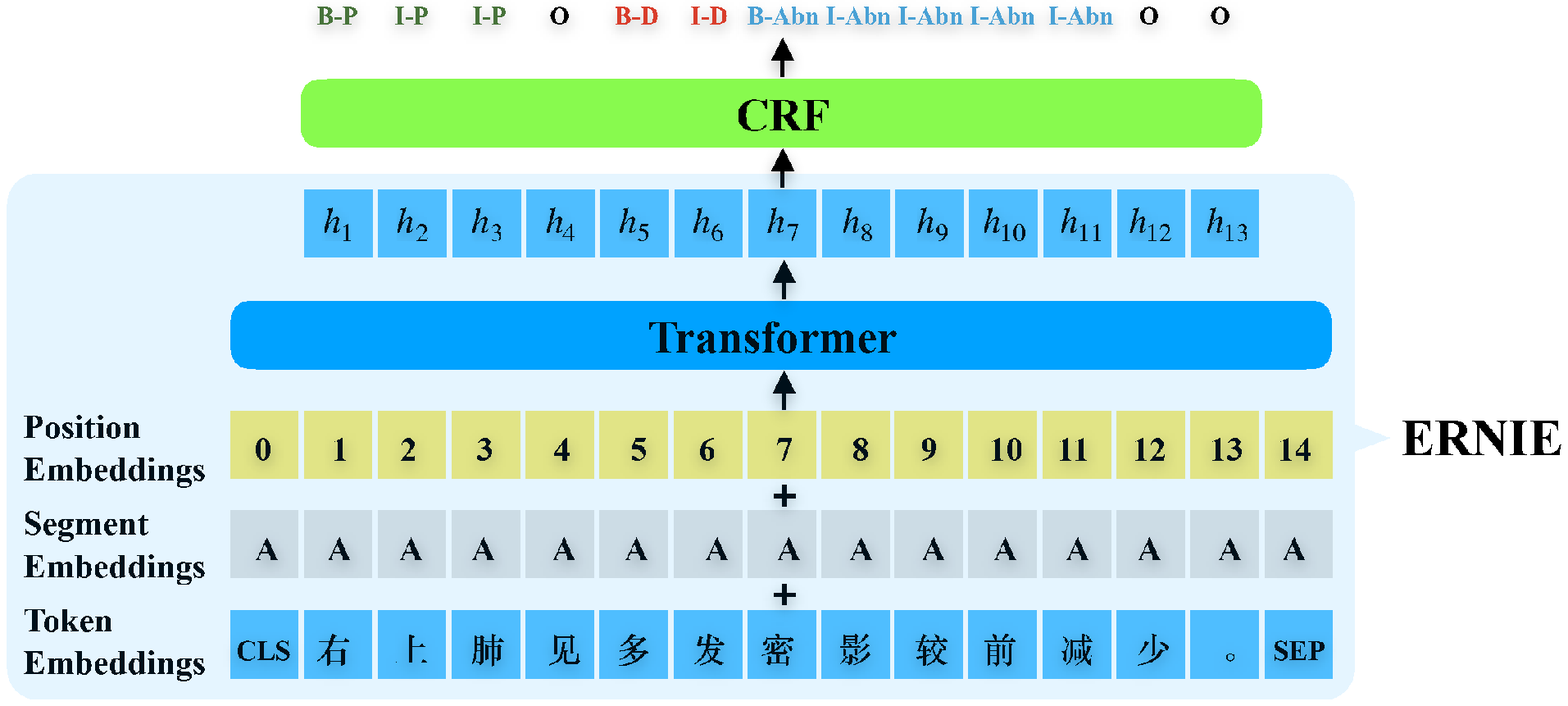}}
\caption{The overview architecture of the EASON model.}
\label{fig3}
\end{figure}

Fig.~\ref{fig3} shows the main structure of our EASON model. We take the input sentence $S=\{x_{i}\}^{n}_{i=1}$ and its corresponding label sequence $y=\{y_{i}\}^{n}_{i=1}$ as an example to introduce each component of EASON from bottom to top as follows, where $n$ is the length of the $S$.

\subsubsection{Encoding sentences with ERNIE}

It is difficult to train a superior deep learning model without any prior knowledge in the case of data insufficiency. In this paper, we use transfer learning to alleviate the problem of data insufficiency. Specifically, we propose to fine-tune ERNIE \cite{DBLP:journals/corr/abs-1904-09223} with CRF for our task, where ERNIE is a novel language representation model based on a multilayer transformer \cite{DBLP:conf/nips/VaswaniSPUJGKP17} and a multistage knowledge masking strategy.

To encode sentence $S$ with ERNIE, we first construct input representations $\bm{H}_{0}$ by summing the corresponding token embeddings ($\bm{E}_{token}$), segment embeddings ($\bm{E}_{seg}$), and position embeddings ($\bm{E}_{pos}$) as follows:

\begin{equation}
\bm{H}_{0}=\bm{E}_{token}+\bm{E}_{seg}+\bm{E}_{pos}
\end{equation}

Then, $N$ transformer layers are applied to calculate the context-dependent representations:

\begin{equation}
\bm{H}_{\alpha}=Transformer(\bm{H}_{\alpha-1}), \alpha \in [1, N],
\end{equation}

where $\bm{H}_{\alpha}$ represents the hidden representations of the input sentence at the $\alpha$-th layer. Note that in addition to fine-tuning, there is another paradigm for transfer learning: feature extraction. In feature extraction, the weights of ERNIE are ``frozen," and the pretrained representations (can also be obtained from the above equation) are used in a downstream model. Section~\ref{sec5.2} shows the relative performance of fine-tuning vs. feature extraction.

Finally, we take the final hidden representations $\bm{H}_{N} \in \mathbb{R}^{n \times d}$ \footnote{Since we focus on intrasentence sequence tagging in this work, we ignore the special classification token CLS and separation token SEP.}, and then project $\bm{H}_{N}$ with a linear projection to matrix $\bm{H}_{N}\bm{W}$, where $d$ is the size of the transformer layer, weight matrix $\bm{W} \in \mathbb{R}^{d \times k}$ is the parameter of the model to be learned in training, and $k$ is the number of distinct tags.

\subsubsection{Conditional Random Field}

The conditional random field (CRF) \cite{DBLP:conf/icml/LaffertyMP01} can obtain a globally optimal chain of labels for a given sequence considering the correlations between adjacent tags. In a sequence tagging task, there are usually strong dependencies between the output labels. Therefore, instead of only fine-tuning ERNIE to model tagging decisions separately, we stack the CRF layer on top of the ERNIE outputs to jointly decode labels for the whole sentence.

We use $\bm{P} \in \mathbb{R}^{n \times k}$ as the matrix of scores output by the linear layer, where $\bm{P}_{ij}$ represents the score of the $j^{th}$ label of the $i^{th}$ character within a sentence. For the sentence $S=\{x_i\}_{i=1}^n$ along with a path of tags $y=\{y_i\}_{i=1}^n$, CRF obtains a real-valued score as follows

\begin{equation}
score\left(S,y\right)=\sum_{i=0}^{n}\bm{A}_{y_{i},y_{i+1}}+\sum_{i=1}^{n}\bm{P}_{i,y_{i}},
\end{equation}

where $\bm{A}$ is the transition matrix, and $\bm{A}_{i,j}$ denotes the score of a transition from tag $i$ to tag $j$. $y_{0}$ and $y_{n}$ are the special tags at the beginning and the end of a sentence, so $\bm{A}$ is a square matrix of size $k+2$. Therefore, the probability for the label sequence $y$ given a sentence $S$ is:

\begin{equation}
p\left(y\mid S\right)=\frac{e^{score\left(S,y\right)}}{\sum_{\widetilde{y}\in Y_{S}}e^{score\left(S,\widetilde{y}\right)}}
\end{equation}

We now maximize the log-likelihood of the correct tag sequence:

\begin{equation}
\log\left(p\left(y\mid S\right)\right)=score\left(S,y\right)-\log\left(\sum_{\widetilde{y}\in Y_{S}}e^{score\left(S,\widetilde{y}\right)}\right),
\end{equation}

where $Y_S$ represents all possible tag sequences for an input sentence $S$. From the formulation above, we can obtain a valid output sequence. When decoding, the sequence with the maximum score is output by:

\begin{equation}
y^{*} = arg\max_{\widetilde{y}\in Y_{S}}score\left(S,\widetilde{y}\right)
\end{equation}

In general, we can use the Viterbi algorithm \cite{DBLP:journals/tit/Viterbi67} to decode the optimal label sequence. Note that the CRF layer is jointly fine-tuned with ERNIE.

\subsection{Matching between abnormal imaging signs and attributes: tag2relation algorithm \label{sec3.4}}

After extracting abnormal imaging signs and attributes, we design a simple but effective matching algorithm \textbf{tag2relation} to assign attributes to the corresponding abnormal imaging signs automatically. Based on the nature of chest imaging text, we find that the secondary body parts of the reports are enumerable, so we asked professional medical practitioners to develop a dictionary of secondary body parts, which was constructed according to the information of secondary body parts in chest imaging reports as well as some medical literature such as 《医学影像学诊断图谱和报告（第一版）》 (\textit{Medical Imaging Diagnostic Atlas and Reports [First Edition]}). To better illustrate this algorithm, we define the semantic unit chunk as follows:

\begin{itemize}
\item \textbf{Chunk:} A chunk refers to the textual content between two primary body parts in the sentence.
\end{itemize}

Furthermore, the relations between attributes and corresponding abnormal imaging signs can be defined as follows:

\begin{itemize}
\item \textbf{P2Abn:} A relation ``P2Abn'' indicates that a primary or secondary body part serves as an attribute of a corresponding abnormal imaging sign
\item \textbf{D2Abn:} A relation ``D2Abn'' indicates that a degree serves as an attribute of a corresponding abnormal imaging sign
\item \textbf{P2P:} A relation ``P2P" indicates that a secondary body part is a subdivision of a primary body part.
\end{itemize}

\IncMargin{1em}
\begin{algorithm}[htbp]
\SetKwData{OutDegree}{out-degree}
\SetKwFunction{Filter}{Filter}\SetKwFunction{Max}{Max}
\SetKwInOut{Input}{input}\SetKwInOut{Output}{output}
\Input{A tag sequence $\mathcal{Y}$ corresponding to sentence $\mathcal{S}$, the dictionary $\mathcal{D}$ of secondary body parts}
\Output{The relations set $\mathcal{T}$ in sentence $\mathcal{S}$}
\BlankLine
\emph{find the primary body parts $\mathcal{P}$ in $\mathcal{S}$ with $\mathcal{Y}$ and $\mathcal{D}$}\;
\emph{find the chunk set $\mathcal{C}$ in $\mathcal{S}$ with $\mathcal{P}$}\;
\emph{$\mathcal{T} \leftarrow \emptyset$}\;
\For{$C \in \mathcal{C}$}{
\emph{find entities $E$ in $C$ with $\mathcal{Y}$}\;
\emph{$R \leftarrow$ perform cartesian products over $E$}\;
\emph{$\mathcal{T} \leftarrow \mathcal{T} \cup $ \Filter{$R$, $\mathcal{S}$, $\mathcal{Y}$}}
}
\KwRet{$\mathcal{T}$}
\caption{Tag2triplet$(\mathcal{S}, \mathcal{Y}, \mathcal{D})$}\label{algo1}
\end{algorithm}\DecMargin{1em}

\begin{figure}[htbp]
\centerline{\includegraphics[scale=0.68]{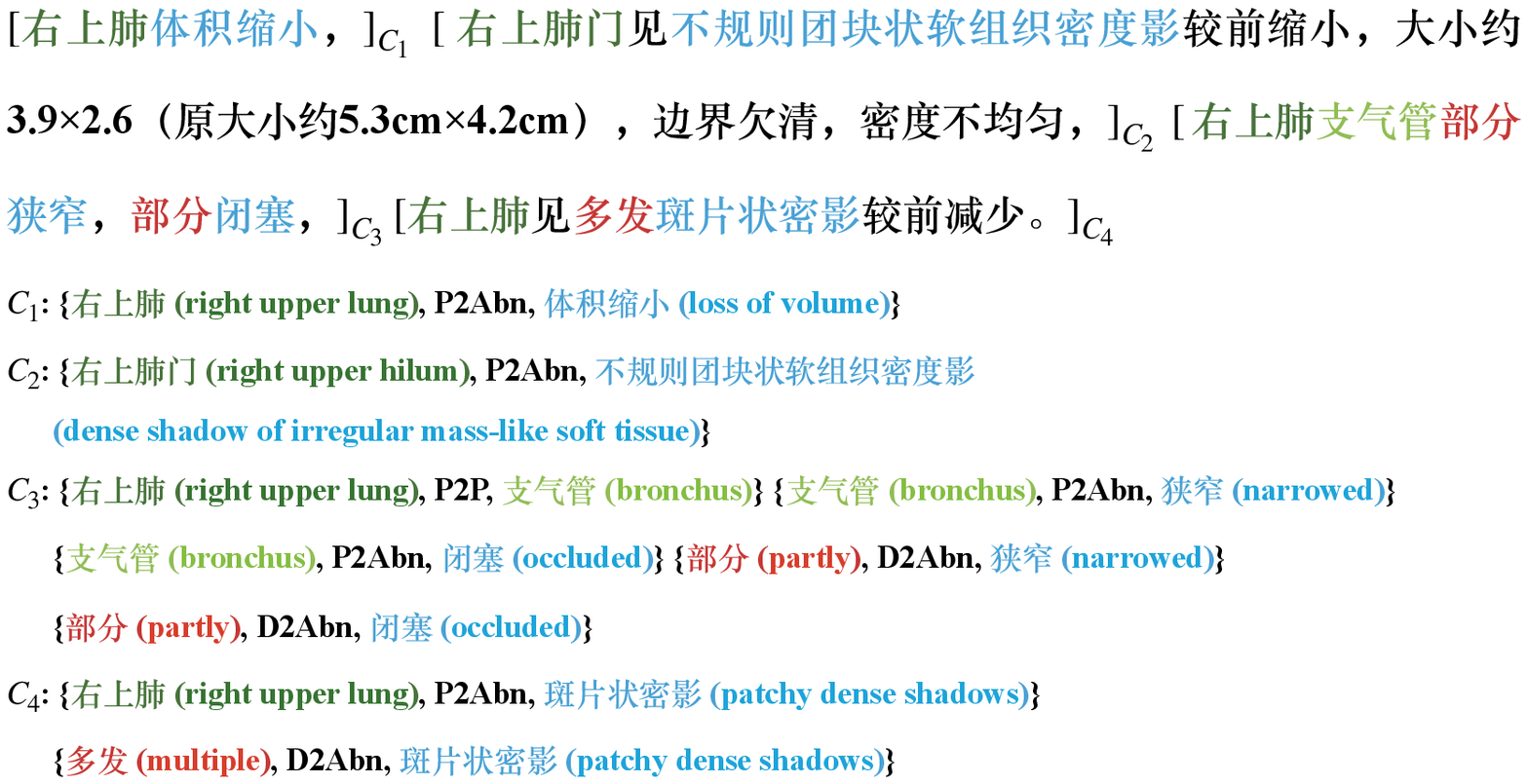}}
\caption{The primary body parts divide the example sentence into four chunks. In this case, $C_{i}$ represents the i-th chunk in the above sentence; \textbf{\textcolor{Cyan}{cyan}} denotes the abnormal imaging signs, \textbf{\textcolor{Red}{red}} denotes the degree of the abnormal imaging sign, \textbf{\textcolor{OliveGreen}{olive green}} and \textbf{\textcolor{LimeGreen}{lime green}} denote the primary and secondary body part of the abnormal imaging sign, respectively.}
\label{fig4}
\end{figure}

The tag2relation algorithm is described in Algorithm~\ref{algo1}. We elaborate on this algorithm by taking the sentence $S$ in Fig.~\ref{fig4} as an example (the English translation of example sentence is shown in Fig.~\ref{fig1}).

First, we use the predefined dictionary to identify the primary body parts (i.e., all the body parts that are not in the dictionary) in the example sentence and then identify each chunk through these parts. As shown in Fig.~\ref{fig4}, we can divide the example sentence into four chunks. In each chunk, we apply a cartesian product over the entity tags to obtain the candidates of the relation. Finally, we \textbf{filter} the candidates by selecting the relations with the shortest distance between attribute and \textit{Abn} to obtain the final matching results.

\section{Experiments}

\subsection{Dataset}

The experimental dataset consists of chest imaging reports provided by a medical big data company, which is a Chinese high-tech enterprise focusing on the construction of a big data management cloud platform for respiratory disease. We asked two annotators with the medical background to manually annotate abnormal imaging signs and corresponding attributes in reports, and the disagreements between two annotators were resolved by a senior medical practitioner. Specifically, our annotation task consists of two subtasks: 1) entity annotation: choosing nonoverlapping entity spans and 2) semantic relation annotation: building a directed graph on top of the entity spans. Based on the annotation results of the two annotators (i.e., A and B), we use F1-score (F1) for consistency evaluation to ensure the quality of the data annotation. The F1 value can be calculated by the following formulas:

\begin{eqnarray}
P&=&\frac{\text{\#identical samples}}{\text{\#total samples labeled by A}},\\
R&=&\frac{\text{\#identical samples}}{\text{\#total samples labeled by B}}, \\
F_{1}&=&2\frac{P\cdot R}{P+R},
\end{eqnarray}

\begin{table}[]
\caption{Annotation statistics}
\begin{center}
\begin{tabular}{ccccc}
\toprule
\multirow{2}{*}{Annotation Process} & \multicolumn{2}{c}{Entity} & \multicolumn{2}{c}{Relation} \\ \cline{2-5} 
                                    & Total     & \textit{F1}    & Total      & \textit{F1}     \\ \hline
First-Round                         & 1398      & 66.25          & 1079       & 60.09           \\
Second-Round                        & 1199      & 85.63          & 885        & 81.84           \\
Official Round                      & 3831      & 93.35          & 3004       & 88.01           \\ \bottomrule
\end{tabular}
\label{tab0}
\end{center}
\end{table}

\begin{table}
\caption{Statistics of abnormal imaging signs and corresponding attributes}
\begin{center}
\begin{tabular}{ccc}
\toprule
Entity Type           & Training Set & Test Set \\ \hline
Abnormal Imaging Sign & 2362         & 428      \\
Body Part             & 2154         & 396      \\
Degree                & 926          & 162      \\ \hline
Sum                   & 5442         & 986      \\ \bottomrule
\end{tabular}
\label{tab1}
\end{center}
\end{table}

The annotation took place in two preannotation rounds and one official annotation round. The purpose of preannotation is to let the annotators fully understand and familiarize themselves with the annotation guidelines, while the senior medical practitioner will further refine the guidelines based on disagreements \footnote{The annotation guidelines are available at https://github.com/Das-Boot/eason.}. Table~\ref{tab0} shows the total number of entities, semantic relations, and consistency F1 values for each round. The F1 values for entities and semantic relation annotation in the second annotation round are 85.63 and 81.84, respectively. When the F1 value is greater than 80\%, we believe that the annotators are already familiar with the annotation guidelines \cite{DBLP:journals/coling/ArtsteinP08}. Then we started the official annotation round. After annotation, we randomly divided the dataset into the training set (253 reports, 2596 sentences) and test set (45 reports, 458 sentences) according to the ratio 0.85:0.15. Table~\ref{tab1} shows the statistics of abnormal imaging signs and their attributes, and Table~\ref{tab2} shows the statistics of relations between abnormal imaging signs and corresponding attributes.

\begin{table}
\caption{Statistics of different types of relations}
\begin{center}
\begin{tabular}{ccc}
\toprule
Relation Type & Training Set & Test Set \\ \hline
P2Abn         & 2721         & 466      \\
D2Abn         & 1126         & 195      \\
P2P           & 388          & 72       \\ \hline
Sum           & 4235         & 733      \\ \bottomrule
\end{tabular}
\label{tab2}
\end{center}
\end{table}

\subsection{Evaluation Metrics}

The standard and widely used performance measures \cite{DBLP:journals/ijmcmc/LiuZWT14,zhou2014correlation}, such as precision (P), recall (R) and F1, are used as evaluation metrics in the following experiments, which can be calculated by the following formulas:

\begin{eqnarray}
P&=&\frac{\text{\#correct predicted samples}}{\text{\#predicted samples}},\\
R&=&\frac{\text{\#correct predicted samples}}{\text{\#total samples in D}}, \\
F_{1}&=&2\frac{P\cdot R}{P+R}, 
\end{eqnarray}

where $D$ is the set of all the sentences in the dataset. In the tasks of abnormal imaging sign ($Abn$) identification and attribute identification ($P$ and $D$), a predicted sample is regarded as correct if and only if it precisely matches an annotated entity. In the task of matching between $Abn$ and attributes, a predicted sample is considered to be correct when its relation type and two corresponding entities are both correct.

\subsection{Hyperparameters}

The model was implemented by using Keras \footnote{https://github.com/keras-team/keras} version 2.2.4 and the ``ERNIE 1.0 Base for Chinese''   \footnote{https://github.com/PaddlePaddle/ERNIE} version of ERNIE in which it uses 12 transformer encoder layers, 768 hidden units, and 12 attention heads. The optimization method of the fine-tuning process was Adam \cite{DBLP:journals/corr/KingmaB14} with $\beta_1=0.9$, $\beta_2=0.999$. The learning rate reached 5e-5 in the first epoch, decayed to 1e-5 in the second epoch, and maintained this learning rate until the end of the training. We let the mini-batch size be 16. In the experiments, we performed the grid search and 10-fold cross-validation on the training set to find the optimal hyperparameters. On the test set, we selected the optimal model among all 200 epochs with the highest validation F1-score.

\subsection{Baselines}

For a comprehensive comparison, we compare our method against several classical sequence tagging models, which can be divided into two categories: CNN-based models and BiLSTM-based models.

For the CNN-based models, the baselines are as follows:

\begin{itemize}
\item \textbf{IDCNN} \cite{DBLP:conf/emnlp/StrubellVBM17}: This model uses a deep iterated dilated CNN (IDCNN) architecture to aggregate context from the entire text, which has better capacity than traditional CNN and faster computation speed than LSTMs and then maps the output of IDCNNs to predict each label independently through a softmax classifier.
\item \textbf{IDCNN-CRF} \cite{DBLP:conf/emnlp/StrubellVBM17}: This model uses CRF to maximize the label probability of the complete sentence based on IDCNNs. Compared to the softmax classifier, the CRF classifier is more appropriate for tasks with strong output label dependency.
\item \textbf{RDCNN-CRF} \cite{DBLP:conf/bibm/QiuWZRG18}: This model is an extension of IDCNN-CRF that uses residual connection \cite{DBLP:conf/cvpr/HeZRS16} between IDCNN layers (RDCNNs) to ease the training of networks and then sums the output of standard CNNs and RDCNNs as the input of CRF.
\end{itemize}

The baselines for the BiLSTM-based models are listed as follows:

\begin{itemize}
\item \textbf{BiLSTM} \cite{DBLP:journals/corr/WangQSHZ15}: The model consists of two parts: a BiLSTM encoder and a softmax classifier.
\item \textbf{BiLSTM-CRF} \cite{DBLP:journals/corr/HuangXY15}: A classic and popular choice for sequence tagging tasks, which consists of a BiLSTM encoder and a CRF classifier.
\end{itemize}

To further analyze the performance of the fine-tuning pretrained Chinese language representation model on our task, we also fine-tune several advanced pretrained models as the experimental baselines:

\begin{itemize}
\item \textbf{BERT} \cite{DBLP:conf/naacl/DevlinCLT19}: BERT (bidirectional encoder representations from the transformer) is the first language representation model based on the bidirectional transformer and masking strategy, and it has shown marvelous improvements across various NLP tasks.
\item \textbf{BERT-wwm} \cite{DBLP:journals/corr/abs-1906-08101}: Bert-wwm is an
upgraded the version of BERT in which they adapted the whole word masking (WWM) strategy in Chinese text for the language model pretraining task.
\item \textbf{ERNIE} \cite{DBLP:journals/corr/abs-1904-09223}: ERNIE is an upgraded version of BERT in which they use phrase-level and entity-level masking strategy in addition to basic masking strategy. Note that ERNIE was trained on not only Chinese Wikipedia data but also Baidu Baike (similar to Wikipedia), Baidu news and Baidu Tieba (similar to Reddit). The numbers of sentences are 21M, 51M, 47M, 54M, respectively.
\end{itemize}

\subsection{Experimental Results \label{sec4.5}}

\begin{table}
\setlength{\tabcolsep}{3.6pt}
\footnotesize
\caption{Comparative results of our EASON model and baseline models on the test set}
\begin{center}
\begin{tabular}{ccccccccccccc}
\toprule
\multirow{2}{*}{Model} & \multicolumn{3}{c}{\textit{Abn} Identification} & \multicolumn{3}{c}{Attributes Identification} & \multicolumn{3}{c}{Matching} \\ \cline{2-10} 
                       & \textit{P}       & \textit{R}       & \textit{F1}     & \textit{P}      & \textit{R}      & \textit{F1}         & \textit{P}      & \textit{R}      & \textit{F1}     \\ \hline
IDCNN                 & 93.93  & 93.93  & 93.93  & 90.58  & 89.61  & 90.09 & 83.36 & 84.38 & 83.87 \\
RDCNN-CRF             & 93.36  & 95.33  & 94.34 & 91.70 & 91.04 & 91.37 & 85.14 & 86.30 & 85.71 \\
IDCNN-CRF             & 95.08  & 94.86  & 94.97 & 92.55 & 91.22 & 91.88  & 86.30 & 86.30 & 86.30 \\ \hline
BiLSTM                 & 94.12  & 93.46  & 93.79 & 91.97 & 90.32 & 91.14  & 84.75 & 84.52 & 84.64 \\
BiLSTM-CRF             & 93.72  & 94.16  & 93.94 & 93.65 & 92.47 & 93.06 & 86.10 & 87.40 & 86.74 \\ \hline
BERT                   & 94.23  & 95.33  & 94.77 & 93.15 & 92.65 & 92.90 & 84.51 & 88.22 & 86.33 \\
BERT-wwm               & 94.21  & 95.09  & 94.65 & 92.83 & 92.83 & 92.83 & 84.94 & 88.08 & 86.48 \\
ERNIE                  & 95.29  & 94.63  & 94.96 & 94.04 & 93.37 & 93.71 & 85.64 & 87.40 & 86.51 \\ \hline
\textbf{EASON}                  & \textbf{96.46}  & \textbf{95.56}  & \textbf{96.01} & \textbf{94.24} & \textbf{93.91} & \textbf{94.08} & \textbf{87.89} & \textbf{89.45} & \textbf{88.66} \\ \bottomrule
\end{tabular}
\label{tab3}
\end{center}
\end{table}

As mentioned in Section~\ref{sec3.1}, our task includes abnormal imaging sign (\textit{Abn}) identification, attribute (\textit{P} and \textit{D}) identification, and matching between \textit{Abn} and attributes. Thus, in this work, we compare our model with baselines in these three subtasks, as shown in Table~\ref{tab3}.

First, we observe that our EASON model outperforms all other models with 96.01\% in \textit{Abn} identification, 94.08\% in attributes identification, and 88.66\% in matching in terms of F1-score. This demonstrates the effectiveness of our proposed method.

Second, compared with superiors of the classical sequence tagging models in F1-score, we can see EASON achieves an improvement of 1.04 points (compared with IDCNN-CRF) in \textit{Abn} identification, 1.02 points (compared with BiLSTM-CRF) in attributes identification, and 1.92 points (compared with BiLSTM-CRF) in matching, which verifies our assumption that the current annotated dataset is not large enough to train a deep learning model sufficiently. With the help of transferred prior knowledge from the pretrained language representation model, we can obtain better performance in all three subtasks.

\begin{figure}
\centerline{\includegraphics[scale=0.52]{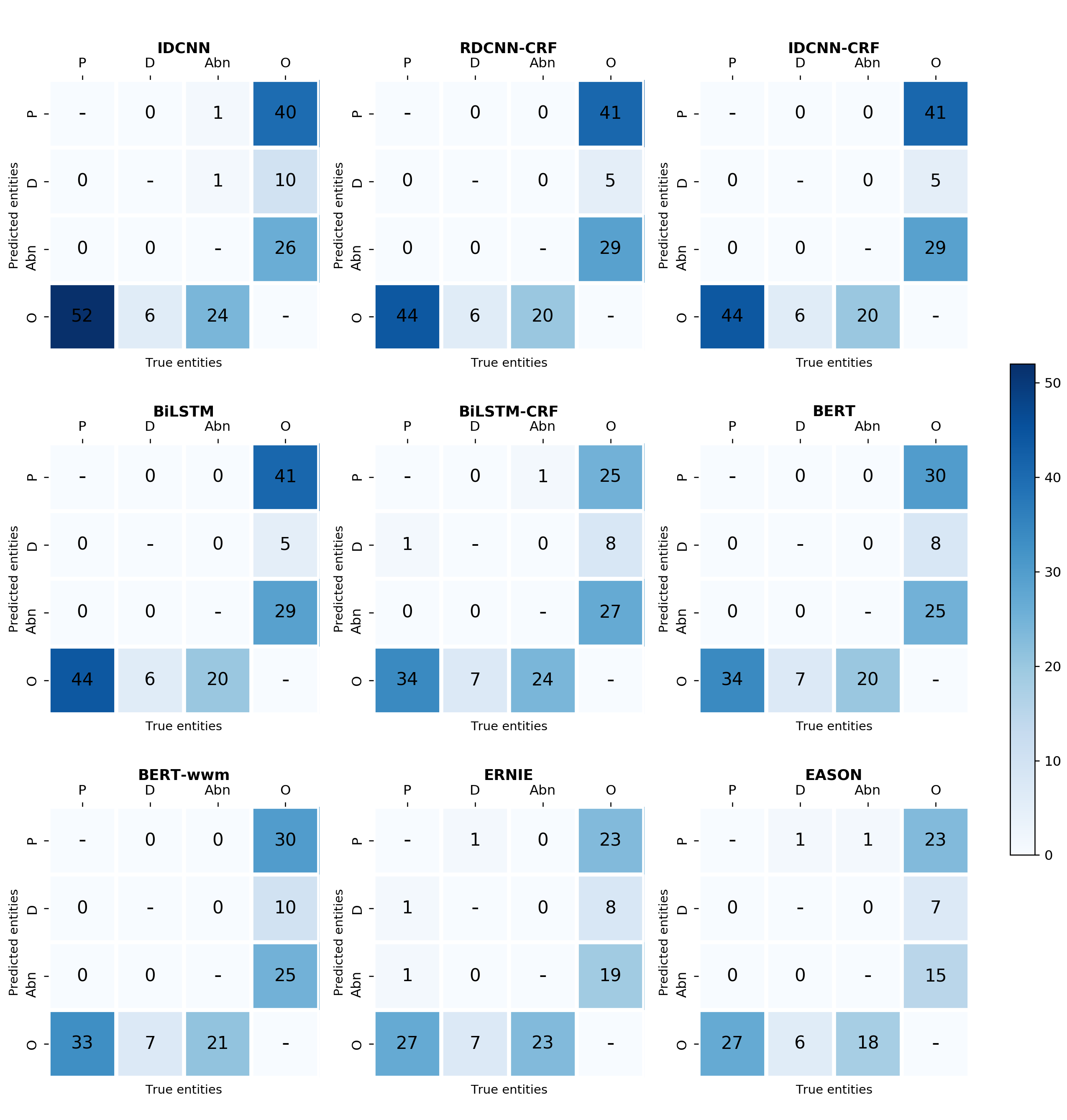}}
\caption{The confusion matrix of our EASON model and other baseline models for entity errors. $X$-Axis: true entities; $y$-axis: predicted entities; P, D, Abn represent body part, degree, abnormal imaging sign entities, respectively, and O is the entities unrelated to the task (i.e., the corresponding character of O is irrelevant in P, D, and Abn).}
\label{fig5}
\end{figure}

Moreover, it shows that the performance of the ERNIE (the output layer is a softmax classifier) improved after jointly fine-tuning with the CRF. The benefits in F1-score brought by jointly fine-tuning with CRF are 1.05, 0.37, 2.15 points in all subtasks because CRF models use the whole label sequence instead of independent label classification and thus can avoid some invalid label sequences.

\section{Analysis and Discussion \label{sec5}}

\subsection{Error Analysis}

\subsubsection{Comparison on Confusion Matrix}

In this paper, we focus on extracting all the quadruples (Section~\ref{sec3.1}) from chest imaging reports, which includes abnormal imaging sign identification, attribute identification, and matching between them. Accurate identification of \textit{Abn}, \textit{P}, and \textit{D} plays a vital role in our task. To visually compare how many errors each model makes at the entity level, we present a confusion matrix for entities \textit{Abn}, \textit{P}, and \textit{D} shown in Fig~\ref{fig5}. We can see that our model EASON (shown in the lower right corner of Fig~\ref{fig5}) can better identify \textit{Abn} and \textit{P} compared with other baselines. Next, we elaborate on the errors produced by EASON.

\subsubsection{Error Analysis of EASON}

To perform error analysis of EASON, we divide the errors under the entity level strict evaluation into four categories according to \cite{DBLP:journals/jamia/WellnerHMAMPYHH07,Jiang_2017}:

\begin{table}
\caption{Statistics of different types of errors produced by EASON}
\begin{center}
\begin{tabular}{ccc}
\toprule
         & Count & \% of Errors \\ \hline
TYPE     & 2     & 2.5\%        \\
EXTENT   & 30    & 37.5\%       \\
SPURIOUS & 20    & 25.0\%       \\
MISSING  & 28    & 35.0\%       \\ \bottomrule
\end{tabular}
\label{tab4}
\end{center}
\end{table}

\begin{itemize}
\item \textbf{Type error}: The entity was identified by a correct span but the incorrect label type.
\item \textbf{Extent error}: The span of the entity overlapped with that of a gold-standard entity but did not match it exactly.
\item \textbf{Spurious error}: The span of the entity had no overlap with any gold-standard entity.
\item \textbf{Missing error}: The span of the entity in the gold standard had no overlap with that of any entity in the system output.
\end{itemize}

\begin{table}
\caption{The distribution of \textbf{type error}, \textcolor{red}{missing error}, and \textcolor{blue}{spurious error}}
\begin{center}
\begin{tabular}{cccccc}
\toprule
\multirow{2}{*}{Gold Standard} & \multicolumn{5}{c}{EASON Output}                            \\ \cline{2-6} 
                  & \textit{P} & \textit{D} & \textit{Abn} & \textcolor{red}{Missing}    & Total \\ \hline
P                 & 369        &            &              & \textcolor{red}{9 (2.4\%)}  & 378   \\
D                 & \textbf{1}          & 155        &              & \textcolor{red}{5 (3.1\%)}  & 161   \\
Abn               & \textbf{1}          &            & 409          & \textcolor{red}{14 (3.3\%)} & 424   \\
\textcolor{blue}{Spurious}          & \textcolor{blue}{9 (2.4\%)}  & \textcolor{blue}{3 (1.9\%)}  & \textcolor{blue}{8 (1.9\%)}    &            & 20    \\
Total             & 380        & 158        & 417          & 28         & 983   \\ \bottomrule
\end{tabular}
\label{tab5}
\end{center}
\end{table}

Table~\ref{tab4} shows the statistics of all errors produced by EASON. Next, we analyze each type of error in detail. Table~\ref{tab5} shows the distribution of type, missing, and spurious errors produced by EASON. Correctly recognized entities can be seen reading down the diagonal in Table~\ref{tab5}. Type errors appear in \textbf{bold}. Type errors are rare, with only two occurrences, and they are all mistagged as body parts. For example, in the sentence ``食管全程扩张，局部较前增著" (The esophagus expands throughout the area, increasing locally), the gold standard has an instance of ``全程" (throughout the area) tagged as degree, while EASON output is body part.

Spurious errors appear in \textcolor{blue}{blue}. From the row of spurious errors, we see that spurious errors occur most often for body part (2.4\% or 9 out of 380) because ordinary body parts had lexical or semantic features similar to those of body parts. However, body parts serve as attributes to the abnormal imaging signs, and ordinary body parts do not. For example, in the sentence ``两肺膨胀良好" (Both lungs are well distended), EASON identifies ``两肺" (both lungs) as body part, though it is not. Missing errors appear in \textcolor{red}{red} and occur most often for the abnormal imaging signal (14 out of 424 tags or 3.3\% missed). For example, EASON does not identify ``食糜及液体潴留" (chyme and fluid retention) and ``胃腔突破食管裂孔" (gastric cavity breaks through esophageal hiatus) as an abnormal imaging signal. The reason may be that EASON requires slightly more training data to learn these less common abnormal imaging signals.

\begin{table}
\caption{Statistics of extent error in EASON output}
\begin{center}
\begin{tabular}{ccccc}
\toprule
                  & \textit{P} & \textit{D} & \textit{Abn} & Total \\ \hline
SHORT      & 5          & 4          & 7            & 16    \\
LONG       & 12         & 0          & 0            & 12    \\
S\&L       & 2          & 0          & 0            & 2     \\
Total      & 19         & 4          & 7            & 30    \\ \bottomrule
\end{tabular}
\label{tab6}
\end{center}
\end{table}

Table~\ref{tab6} shows statistics for three types of extent errors. Numbers in the short row indicate instances where the span of the entity produced by EASON fell within that of a gold standard. In contrast, numbers in the long row indicate instances where the span of the entity covered that of a gold standard. Moreover, numbers in the short and long row indicate the span of the entity neither fell into nor covered that of a gold standard. Nearly two-thirds of the extent errors occurred on body part, with many instances of long errors. The occurrence of extent error indicates that EASON sometimes cannot detect the boundaries of entities correctly. There are two main reasons for these errors. The first cause is tokenization. For example, the gold standard has an instance of ``肝" (liver), while EASON tagged ``肝、" as body part (long extent error) because of the failure of tokenizing ``、" from ``肝." Another cause is that EASON tags one entity as two (short extent error) or tags two entities as one (long extent error). For example, the body part ``食管下端贲门区" (the lower esophagus from the cardia) identified by EASON is, in fact, two entities in the gold standard: ``食管下端" (lower esophagus) and ``贲门区" (cardia area).

\subsection{Feature extraction or fine-tuning? \label{sec5.2}}

\begin{table}[htbp]
\caption{Test set performance of our EASON model and feature-based models}
\begin{center}
\begin{tabular}{ccccc}
\toprule
\multirow{2}{*}{Model}  & \multicolumn{3}{c}{\textit{F1}}                                         \\ \cline{2-4} 
                        & \textit{Abn}    & Attributes   & Matching                     \\ \hline
First Layer (Embeddings) & 94.55  & 92.46              & 86.39                   \\               
Second-to-Last Hidden   & 94.68  & \textbf{94.28}            & 86.84                     \\
Last Hidden             & 94.83  & 94.03              & 87.14                     \\
Sum Last Four Hidden    & 94.50  & 94.07        & 87.48                     \\ 
Concat Last Four Hidden & 94.91  & 93.48          & 85.98                   \\
Sum All 12 Layers       & 95.25  & 93.92          & 87.02                     \\ \hline
\textbf{EASON}                   & \textbf{96.01}       & 94.08       & \textbf{88.66}                     \\ \bottomrule
\end{tabular}
\end{center}
\label{tab7}
\end{table}

To further investigate the performance of feature extraction (where the weights of the pretrained model are frozen) and fine-tuning the pretrained model in our task, we compare our model with feature-based models that use one or more layers of ERNIE as input to a one-layer 512-dimensional BiLSTM before the CRF layer. The results are illustrated in Table~\ref{tab7}. From the table, we see that EASON achieves No. 1 in \textit{Abn} identification and matching and No. 2 in attributes identification in terms of F1-score. This indicates that our proposed fine-tuning model is more suitable for the task of chest abnormal imaging sign extraction than feature-based models.

\section{Conclusion}

In this paper, we formulate chest abnormal imaging sign extraction as a sequence tagging and matching problem and deliver an effective solution for this task. In particular, we propose EASON to extract abnormal imaging signs and their attributes in Chinese chest imaging reports. To alleviate the problem of data insufficiency, we fine-tune ERNIE trained from the large corpus with CRF to perform sequence tagging in our task. In addition, we design a tag2relation algorithm to assign the attributes to corresponding abnormal imaging signs from the results of the sequence tagging model. Experimental results on the corpus provided by a medical big data company show that our proposed EASON model achieves superior performance compared to other baseline models, i.e., reaching the F1-score of 96.01\%, 94.08\%, 88.66\% in \textit{Abn} identification, attributes identification and matching, respectively.

\section{Acknowledgment}

This work was partially supported by the National Key R\&D Plan of China (Grant No. 2018YFC1315402).

\section*{References}

\bibliography{reference}

\begin{thebibliography}{49}
\providecommand{\natexlab}[1]{#1}
\providecommand{\url}[1]{\texttt{#1}}
\providecommand{\urlprefix}{URL }
\expandafter\ifx\csname urlstyle\endcsname\relax
  \providecommand{\doi}[1]{doi:\discretionary{}{}{}#1}\else
  \providecommand{\doi}[1]{doi:\discretionary{}{}{}\begingroup
  \urlstyle{rm}\url{#1}\endgroup}\fi
\providecommand{\bibinfo}[2]{#2}

\bibitem[{Litjens et~al.(2017)Litjens, Kooi, Bejnordi, Setio, Ciompi,
  Ghafoorian, van~der Laak, van Ginneken, and
  S{\'{a}}nchez}]{DBLP:journals/mia/LitjensKBSCGLGS17}
\bibinfo{author}{G.~J.~S. Litjens}, \bibinfo{author}{T.~Kooi},
  \bibinfo{author}{B.~E. Bejnordi}, \bibinfo{author}{A.~A.~A. Setio},
  \bibinfo{author}{F.~Ciompi}, \bibinfo{author}{M.~Ghafoorian},
  \bibinfo{author}{J.~A. W.~M. van~der Laak}, \bibinfo{author}{B.~van
  Ginneken}, \bibinfo{author}{C.~I. S{\'{a}}nchez}, \bibinfo{title}{A survey on
  deep learning in medical image analysis}, \bibinfo{journal}{Medical Image
  Analysis} \bibinfo{volume}{42} (\bibinfo{year}{2017})
  \bibinfo{pages}{60--88}.

\bibitem[{Lundervold and Lundervold(2019)}]{DBLP:journals/corr/abs-1811-10052}
\bibinfo{author}{A.~S. Lundervold}, \bibinfo{author}{A.~Lundervold},
  \bibinfo{title}{An overview of deep learning in medical imaging focusing on
  MRI}, \bibinfo{journal}{Z Med Phys}
  \bibinfo{volume}{29}~(\bibinfo{number}{2}) (\bibinfo{year}{2019})
  \bibinfo{pages}{102--127}.

\bibitem[{Pons et~al.(2016)Pons, Braun, Hunink, and Kors}]{Pons_2016}
\bibinfo{author}{E.~Pons}, \bibinfo{author}{L.~M.~M. Braun},
  \bibinfo{author}{M.~G.~M. Hunink}, \bibinfo{author}{J.~A. Kors},
  \bibinfo{title}{Natural Language Processing in Radiology: A Systematic
  Review}, \bibinfo{journal}{Radiology}
  \bibinfo{volume}{279}~(\bibinfo{number}{2}) (\bibinfo{year}{2016})
  \bibinfo{pages}{329--343}.

\bibitem[{Ni et~al.(2017)Ni, Liu, Zhang, Ye, and Ma}]{DBLP:conf/cikm/Ni0ZYM17}
\bibinfo{author}{J.~Ni}, \bibinfo{author}{J.~Liu}, \bibinfo{author}{C.~Zhang},
  \bibinfo{author}{D.~Ye}, \bibinfo{author}{Z.~Ma},
  \bibinfo{title}{Fine-grained patient similarity measuring using deep metric
  learning}, in: \bibinfo{booktitle}{Proceedings of the 2017 {ACM} on
  Conference on Information and Knowledge Management, {CIKM}},
  \bibinfo{pages}{1189--1198}, \bibinfo{year}{2017}.

\bibitem[{Mullenbach et~al.(2018)Mullenbach, Wiegreffe, Duke, Sun, and
  Eisenstein}]{DBLP:conf/naacl/MullenbachWDSE18}
\bibinfo{author}{J.~Mullenbach}, \bibinfo{author}{S.~Wiegreffe},
  \bibinfo{author}{J.~Duke}, \bibinfo{author}{J.~Sun},
  \bibinfo{author}{J.~Eisenstein}, \bibinfo{title}{Explainable prediction of
  medical codes from clinical text}, in: \bibinfo{booktitle}{Proceedings of the
  2018 Conference of the North American Chapter of the Association for
  Computational Linguistics: Human Language Technologies, {NAACL-HLT}},
  \bibinfo{pages}{1101--1111}, \bibinfo{year}{2018}.

\bibitem[{Friedman et~al.(1995)Friedman, Hripcsak, DuMouchel, Johnson, and
  Clayton}]{Friedman_1995}
\bibinfo{author}{C.~Friedman}, \bibinfo{author}{G.~Hripcsak},
  \bibinfo{author}{W.~DuMouchel}, \bibinfo{author}{S.~B. Johnson},
  \bibinfo{author}{P.~D. Clayton}, \bibinfo{title}{Natural language processing
  in an operational clinical information system}, \bibinfo{journal}{Natural
  Language Engineering} \bibinfo{volume}{1}~(\bibinfo{number}{1})
  (\bibinfo{year}{1995}) \bibinfo{pages}{83--108}.

\bibitem[{Johnson et~al.(1997)Johnson, Taira, Cardenas, and
  Aberle}]{Johnson_1997}
\bibinfo{author}{D.~B. Johnson}, \bibinfo{author}{R.~K. Taira},
  \bibinfo{author}{A.~F. Cardenas}, \bibinfo{author}{D.~R. Aberle},
  \bibinfo{title}{Extracting information from free text radiology reports},
  \bibinfo{journal}{International Journal on Digital Libraries}
  \bibinfo{volume}{1}~(\bibinfo{number}{3}) (\bibinfo{year}{1997})
  \bibinfo{pages}{297--308}.

\bibitem[{Esuli et~al.(2013)Esuli, Marcheggiani, and
  Sebastiani}]{DBLP:journals/jbi/EsuliMS13}
\bibinfo{author}{A.~Esuli}, \bibinfo{author}{D.~Marcheggiani},
  \bibinfo{author}{F.~Sebastiani}, \bibinfo{title}{An enhanced CRFs-based
  system for information extraction from radiology reports},
  \bibinfo{journal}{Journal of Biomedical Informatics}
  \bibinfo{volume}{46}~(\bibinfo{number}{3}) (\bibinfo{year}{2013})
  \bibinfo{pages}{425--435}.

\bibitem[{Bozkurt et~al.(2015)Bozkurt, Lipson, Senol, and
  Rubin}]{DBLP:journals/jamia/BozkurtLSR15}
\bibinfo{author}{S.~Bozkurt}, \bibinfo{author}{J.~A. Lipson},
  \bibinfo{author}{U.~Senol}, \bibinfo{author}{D.~L. Rubin},
  \bibinfo{title}{Automatic abstraction of imaging observations with their
  characteristics from mammography reports}, \bibinfo{journal}{{JAMIA}}
  \bibinfo{volume}{22}~(\bibinfo{number}{e1}) (\bibinfo{year}{2015})
  \bibinfo{pages}{e81--e92}.

\bibitem[{Hassanpour and Langlotz(2016)}]{DBLP:journals/artmed/HassanpourL16}
\bibinfo{author}{S.~Hassanpour}, \bibinfo{author}{C.~P. Langlotz},
  \bibinfo{title}{Information extraction from multi-institutional radiology
  reports}, \bibinfo{journal}{Artificial Intelligence in Medicine}
  \bibinfo{volume}{66} (\bibinfo{year}{2016}) \bibinfo{pages}{29--39}.

\bibitem[{Gupta et~al.(2018)Gupta, Banerjee, and
  Rubin}]{DBLP:journals/jbi/GuptaBR18}
\bibinfo{author}{A.~Gupta}, \bibinfo{author}{I.~Banerjee},
  \bibinfo{author}{D.~L. Rubin}, \bibinfo{title}{Automatic information
  extraction from unstructured mammography reports using distributed
  semantics}, \bibinfo{journal}{Journal of Biomedical Informatics}
  \bibinfo{volume}{78} (\bibinfo{year}{2018}) \bibinfo{pages}{78--86}.

\bibitem[{McCallum et~al.(2000)McCallum, Freitag, and
  Pereira}]{DBLP:conf/icml/McCallumFP00}
\bibinfo{author}{A.~McCallum}, \bibinfo{author}{D.~Freitag},
  \bibinfo{author}{F.~C.~N. Pereira}, \bibinfo{title}{Maximum entropy Markov
  models for information extraction and segmentation}, in:
  \bibinfo{booktitle}{Proceedings of the Seventeenth International Conference
  on Machine Learning, {ICML}}, \bibinfo{pages}{591--598},
  \bibinfo{year}{2000}.

\bibitem[{Zhou and Su(2002)}]{DBLP:conf/acl/ZhouS02}
\bibinfo{author}{G.~Zhou}, \bibinfo{author}{J.~Su}, \bibinfo{title}{Named
  entity recognition using an HMM-based chunk tagger}, in:
  \bibinfo{booktitle}{Proceedings of the 40th Annual Meeting of the Association
  for Computational Linguistics, {ACL}}, \bibinfo{pages}{473--480},
  \bibinfo{year}{2002}.

\bibitem[{McCallum and Li(2003)}]{DBLP:conf/conll/McCallum003}
\bibinfo{author}{A.~McCallum}, \bibinfo{author}{W.~Li}, \bibinfo{title}{Early
  results for named entity recognition with conditional random fields, feature
  induction and web-enhanced lexicons}, in: \bibinfo{booktitle}{Proceedings of
  the Seventh Conference on Natural Language Learning, {CoNLL}},
  \bibinfo{pages}{188--191}, \bibinfo{year}{2003}.

\bibitem[{Hochreiter and Schmidhuber(1997)}]{DBLP:journals/neco/HochreiterS97}
\bibinfo{author}{S.~Hochreiter}, \bibinfo{author}{J.~Schmidhuber},
  \bibinfo{title}{Long short-term memory}, \bibinfo{journal}{Neural
  Computation} \bibinfo{volume}{9}~(\bibinfo{number}{8}) (\bibinfo{year}{1997})
  \bibinfo{pages}{1735--1780}.

\bibitem[{LeCun et~al.(1989)LeCun, Boser, Denker, Henderson, Howard, Hubbard,
  and Jackel}]{DBLP:journals/neco/LeCunBDHHHJ89}
\bibinfo{author}{Y.~LeCun}, \bibinfo{author}{B.~E. Boser},
  \bibinfo{author}{J.~S. Denker}, \bibinfo{author}{D.~Henderson},
  \bibinfo{author}{R.~E. Howard}, \bibinfo{author}{W.~E. Hubbard},
  \bibinfo{author}{L.~D. Jackel}, \bibinfo{title}{Backpropagation applied to
  handwritten zip code recognition}, \bibinfo{journal}{Neural Computation}
  \bibinfo{volume}{1}~(\bibinfo{number}{4}) (\bibinfo{year}{1989})
  \bibinfo{pages}{541--551}.

\bibitem[{Lafferty et~al.(2001)Lafferty, McCallum, and
  Pereira}]{DBLP:conf/icml/LaffertyMP01}
\bibinfo{author}{J.~D. Lafferty}, \bibinfo{author}{A.~McCallum},
  \bibinfo{author}{F.~C.~N. Pereira}, \bibinfo{title}{Conditional random
  fields: probabilistic models for segmenting and labeling sequence data}, in:
  \bibinfo{booktitle}{Proceedings of the Eighteenth International Conference on
  Machine Learning, {ICML}}, \bibinfo{pages}{282--289}, \bibinfo{year}{2001}.

\bibitem[{Habibi et~al.(2017)Habibi, Weber, Neves, Wiegandt, and
  Leser}]{DBLP:journals/bioinformatics/HabibiWNWL17}
\bibinfo{author}{M.~Habibi}, \bibinfo{author}{L.~Weber}, \bibinfo{author}{M.~L.
  Neves}, \bibinfo{author}{D.~L. Wiegandt}, \bibinfo{author}{U.~Leser},
  \bibinfo{title}{Deep learning with word embeddings improves biomedical named
  entity recognition}, \bibinfo{journal}{Bioinformatics}
  \bibinfo{volume}{33}~(\bibinfo{number}{14}) (\bibinfo{year}{2017})
  \bibinfo{pages}{i37--i48}.

\bibitem[{Wang et~al.(2019)Wang, Zhou, Ruan, Gao, Xia, and
  He}]{DBLP:journals/jbi/WangZRGXH19}
\bibinfo{author}{Q.~Wang}, \bibinfo{author}{Y.~Zhou},
  \bibinfo{author}{T.~Ruan}, \bibinfo{author}{D.~Gao},
  \bibinfo{author}{Y.~Xia}, \bibinfo{author}{P.~He},
  \bibinfo{title}{Incorporating dictionaries into deep neural networks for the
  Chinese clinical named entity recognition}, \bibinfo{journal}{Journal of
  Biomedical Informatics} \bibinfo{volume}{92}.

\bibitem[{Qiu et~al.(2018)Qiu, Wang, Zhou, Ruan, and
  Gao}]{DBLP:conf/bibm/QiuWZRG18}
\bibinfo{author}{J.~Qiu}, \bibinfo{author}{Q.~Wang}, \bibinfo{author}{Y.~Zhou},
  \bibinfo{author}{T.~Ruan}, \bibinfo{author}{J.~Gao}, \bibinfo{title}{Fast and
  accurate recognition of Chinese clinical named entities with residual dilated
  convolutions}, in: \bibinfo{booktitle}{{IEEE} International Conference on
  Bioinformatics and Biomedicine, {BIBM}}, \bibinfo{pages}{935--942},
  \bibinfo{year}{2018}.

\bibitem[{Zheng et~al.(2017)Zheng, Wang, Bao, Hao, Zhou, and
  Xu}]{DBLP:conf/acl/ZhengWBHZX17}
\bibinfo{author}{S.~Zheng}, \bibinfo{author}{F.~Wang},
  \bibinfo{author}{H.~Bao}, \bibinfo{author}{Y.~Hao},
  \bibinfo{author}{P.~Zhou}, \bibinfo{author}{B.~Xu}, \bibinfo{title}{Joint
  Extraction of Entities and Relations Based on a Novel Tagging Scheme}, in:
  \bibinfo{booktitle}{Proceedings of the 55th Annual Meeting of the Association
  for Computational Linguistics, {ACL}}, \bibinfo{pages}{1227--1236},
  \bibinfo{year}{2017}.

\bibitem[{Sun et~al.(2019{\natexlab{a}})Sun, Wang, Li, Feng, Chen, Zhang, Tian,
  Zhu, Tian, and Wu}]{DBLP:journals/corr/abs-1904-09223}
\bibinfo{author}{Y.~Sun}, \bibinfo{author}{S.~Wang}, \bibinfo{author}{Y.~Li},
  \bibinfo{author}{S.~Feng}, \bibinfo{author}{X.~Chen},
  \bibinfo{author}{H.~Zhang}, \bibinfo{author}{X.~Tian},
  \bibinfo{author}{D.~Zhu}, \bibinfo{author}{H.~Tian}, \bibinfo{author}{H.~Wu},
  \bibinfo{title}{ERNIE: Enhanced representation through knowledge
  integration}, \bibinfo{journal}{arXiv preprint}
  \bibinfo{volume}{arXiv:1904.09223}.

\bibitem[{Friedman et~al.(1994)Friedman, Alderson, Austin, Cimino, and
  Johnson}]{DBLP:journals/jamia/FriedmanAACJ94}
\bibinfo{author}{C.~Friedman}, \bibinfo{author}{P.~O. Alderson},
  \bibinfo{author}{J.~H.~M. Austin}, \bibinfo{author}{J.~J. Cimino},
  \bibinfo{author}{S.~B. Johnson}, \bibinfo{title}{Research Paper: {A} General
  Natural-language Text Processor for Clinical Radiology},
  \bibinfo{journal}{{JAMIA}} \bibinfo{volume}{1}~(\bibinfo{number}{2})
  (\bibinfo{year}{1994}) \bibinfo{pages}{161--174}.

\bibitem[{Zeng et~al.(2006)Zeng, Goryachev, Weiss, Sordo, Murphy, and
  Lazarus}]{DBLP:journals/midm/ZengGWSML06}
\bibinfo{author}{Q.~T. Zeng}, \bibinfo{author}{S.~Goryachev},
  \bibinfo{author}{S.~T. Weiss}, \bibinfo{author}{M.~Sordo},
  \bibinfo{author}{S.~N. Murphy}, \bibinfo{author}{R.~Lazarus},
  \bibinfo{title}{Extracting principal diagnosis, co-morbidity and smoking
  status for asthma research: evaluation of a natural language processing
  system}, \bibinfo{journal}{{BMC} Med. Inf. {\&} Decision Making}
  \bibinfo{volume}{6} (\bibinfo{year}{2006}) \bibinfo{pages}{30}.

\bibitem[{Coden et~al.(2009)Coden, Savova, Sominsky, Tanenblatt, Masanz,
  Schuler, Cooper, Guan, and de~Groen}]{DBLP:journals/jbi/CodenSSTMSCGG09}
\bibinfo{author}{A.~Coden}, \bibinfo{author}{G.~K. Savova},
  \bibinfo{author}{I.~L. Sominsky}, \bibinfo{author}{M.~A. Tanenblatt},
  \bibinfo{author}{J.~J. Masanz}, \bibinfo{author}{K.~Schuler},
  \bibinfo{author}{J.~W. Cooper}, \bibinfo{author}{W.~Guan},
  \bibinfo{author}{P.~C. de~Groen}, \bibinfo{title}{Automatically extracting
  cancer disease characteristics from pathology reports into a Disease
  Knowledge Representation Model}, \bibinfo{journal}{Journal of Biomedical
  Informatics} \bibinfo{volume}{42}~(\bibinfo{number}{5})
  (\bibinfo{year}{2009}) \bibinfo{pages}{937--949}.

\bibitem[{Harkema et~al.(2009)Harkema, Dowling, Thornblade, and
  Chapman}]{DBLP:journals/jbi/HarkemaDTC09}
\bibinfo{author}{H.~Harkema}, \bibinfo{author}{J.~N. Dowling},
  \bibinfo{author}{T.~Thornblade}, \bibinfo{author}{W.~W. Chapman},
  \bibinfo{title}{ConText: An algorithm for determining negation, experiencer,
  and temporal status from clinical reports}, \bibinfo{journal}{Journal of
  Biomedical Informatics} \bibinfo{volume}{42}~(\bibinfo{number}{5})
  (\bibinfo{year}{2009}) \bibinfo{pages}{839--851}.

\bibitem[{Chapman et~al.(2001)Chapman, Bridewell, Hanbury, Cooper, and
  Buchanan}]{Chapman_2001}
\bibinfo{author}{W.~W. Chapman}, \bibinfo{author}{W.~Bridewell},
  \bibinfo{author}{P.~Hanbury}, \bibinfo{author}{G.~F. Cooper},
  \bibinfo{author}{B.~G. Buchanan}, \bibinfo{title}{A Simple Algorithm for
  Identifying Negated Findings and Diseases in Discharge Summaries},
  \bibinfo{journal}{Journal of Biomedical Informatics}
  \bibinfo{volume}{34}~(\bibinfo{number}{5}) (\bibinfo{year}{2001})
  \bibinfo{pages}{301--310}.

\bibitem[{Song et~al.(2015)Song, Yu, and Han}]{DBLP:journals/midm/SongYH15}
\bibinfo{author}{M.~Song}, \bibinfo{author}{H.~Yu}, \bibinfo{author}{W.~Han},
  \bibinfo{title}{Developing a hybrid dictionary-based bio-entity recognition
  technique}, \bibinfo{journal}{{BMC} Med. Inf. {\&} Decision Making}
  \bibinfo{volume}{15}~(\bibinfo{number}{{S-1}}) (\bibinfo{year}{2015})
  \bibinfo{pages}{S9}.

\bibitem[{Finkel et~al.(2004)Finkel, Dingare, Nguyen, Nissim, Manning, and
  Sinclair}]{DBLP:conf/bionlp/FinkelDNNMS04}
\bibinfo{author}{J.~R. Finkel}, \bibinfo{author}{S.~Dingare},
  \bibinfo{author}{H.~Nguyen}, \bibinfo{author}{M.~Nissim},
  \bibinfo{author}{C.~D. Manning}, \bibinfo{author}{G.~Sinclair},
  \bibinfo{title}{Exploiting context for biomedical entity recognition: from
  syntax to the web}, in: \bibinfo{booktitle}{Proceedings of the International
  Joint Workshop on Natural Language Processing in Biomedicine and its
  Applications, NLPBA/BioNLP}, \bibinfo{year}{2004}.

\bibitem[{Skeppstedt et~al.(2014)Skeppstedt, Kvist, Nilsson, and
  Dalianis}]{DBLP:journals/jbi/SkeppstedtKND14}
\bibinfo{author}{M.~Skeppstedt}, \bibinfo{author}{M.~Kvist},
  \bibinfo{author}{G.~H. Nilsson}, \bibinfo{author}{H.~Dalianis},
  \bibinfo{title}{Automatic recognition of disorders, findings, pharmaceuticals
  and body structures from clinical text: An annotation and machine learning
  study}, \bibinfo{journal}{Journal of Biomedical Informatics}
  \bibinfo{volume}{49} (\bibinfo{year}{2014}) \bibinfo{pages}{148--158}.

\bibitem[{Wu et~al.(2006)Wu, Fan, Lee, and Yen}]{DBLP:conf/pakdd/WuFLY06}
\bibinfo{author}{Y.~Wu}, \bibinfo{author}{T.~Fan}, \bibinfo{author}{Y.~Lee},
  \bibinfo{author}{S.~Yen}, \bibinfo{title}{Extracting named entities using
  support vector machines}, in: \bibinfo{booktitle}{Knowledge Discovery in Life
  Science Literature, {PAKDD} 2006 International Workshop, {KDLL}},
  \bibinfo{pages}{91--103}, \bibinfo{year}{2006}.

\bibitem[{Ju et~al.(2011)Ju, Wang, and Zhu}]{ju2011named}
\bibinfo{author}{Z.~Ju}, \bibinfo{author}{J.~Wang}, \bibinfo{author}{F.~Zhu},
  \bibinfo{title}{Named entity recognition from biomedical text using SVM}, in:
  \bibinfo{booktitle}{International Conference on Bioinformatics and Biomedical
  Engineering}, \bibinfo{pages}{1--4}, \bibinfo{year}{2011}.

\bibitem[{Peters et~al.(2018)Peters, Neumann, Iyyer, Gardner, Clark, Lee, and
  Zettlemoyer}]{DBLP:conf/naacl/PetersNIGCLZ18}
\bibinfo{author}{M.~E. Peters}, \bibinfo{author}{M.~Neumann},
  \bibinfo{author}{M.~Iyyer}, \bibinfo{author}{M.~Gardner},
  \bibinfo{author}{C.~Clark}, \bibinfo{author}{K.~Lee},
  \bibinfo{author}{L.~Zettlemoyer}, \bibinfo{title}{Deep contextualized word
  representations}, in: \bibinfo{booktitle}{Proceedings of the 2018 Conference
  of the North American Chapter of the Association for Computational
  Linguistics: Human Language Technologies, {NAACL-HLT}},
  \bibinfo{pages}{2227--2237}, \bibinfo{year}{2018}.

\bibitem[{Akbik et~al.(2018)Akbik, Blythe, and
  Vollgraf}]{DBLP:conf/coling/AkbikBV18}
\bibinfo{author}{A.~Akbik}, \bibinfo{author}{D.~Blythe},
  \bibinfo{author}{R.~Vollgraf}, \bibinfo{title}{Contextual string embeddings
  for sequence labeling}, in: \bibinfo{booktitle}{Proceedings of the 27th
  International Conference on Computational Linguistics, {COLING}},
  \bibinfo{pages}{1638--1649}, \bibinfo{year}{2018}.

\bibitem[{Devlin et~al.(2019)Devlin, Chang, Lee, and
  Toutanova}]{DBLP:conf/naacl/DevlinCLT19}
\bibinfo{author}{J.~Devlin}, \bibinfo{author}{M.~Chang},
  \bibinfo{author}{K.~Lee}, \bibinfo{author}{K.~Toutanova},
  \bibinfo{title}{{BERT:} Pre-training of deep bidirectional transformers for
  language understanding}, in: \bibinfo{booktitle}{Proceedings of the 2019
  Conference of the North American Chapter of the Association for Computational
  Linguistics: Human Language Technologies, {NAACL-HLT}},
  \bibinfo{pages}{4171--4186}, \bibinfo{year}{2019}.

\bibitem[{Cui et~al.(2019)Cui, Che, Liu, Qin, Yang, Wang, and
  Hu}]{DBLP:journals/corr/abs-1906-08101}
\bibinfo{author}{Y.~Cui}, \bibinfo{author}{W.~Che}, \bibinfo{author}{T.~Liu},
  \bibinfo{author}{B.~Qin}, \bibinfo{author}{Z.~Yang},
  \bibinfo{author}{S.~Wang}, \bibinfo{author}{G.~Hu},
  \bibinfo{title}{Pre-training with whole word masking for Chinese BERT},
  \bibinfo{journal}{arXiv preprint} \bibinfo{volume}{arXiv:1906.08101}.

\bibitem[{Sun et~al.(2019{\natexlab{b}})Sun, Wang, Li, Feng, Tian, Wu, and
  Wang}]{DBLP:journals/corr/abs-1907-12412}
\bibinfo{author}{Y.~Sun}, \bibinfo{author}{S.~Wang}, \bibinfo{author}{Y.~Li},
  \bibinfo{author}{S.~Feng}, \bibinfo{author}{H.~Tian},
  \bibinfo{author}{H.~Wu}, \bibinfo{author}{H.~Wang}, \bibinfo{title}{Ernie
  2.0: A continual pre-training framework for language understanding},
  \bibinfo{journal}{arXiv preprint} \bibinfo{volume}{arXiv:1907.12412}.

\bibitem[{Vaswani et~al.(2017)Vaswani, Shazeer, Parmar, Uszkoreit, Jones,
  Gomez, Kaiser, and Polosukhin}]{DBLP:conf/nips/VaswaniSPUJGKP17}
\bibinfo{author}{A.~Vaswani}, \bibinfo{author}{N.~Shazeer},
  \bibinfo{author}{N.~Parmar}, \bibinfo{author}{J.~Uszkoreit},
  \bibinfo{author}{L.~Jones}, \bibinfo{author}{A.~N. Gomez},
  \bibinfo{author}{L.~Kaiser}, \bibinfo{author}{I.~Polosukhin},
  \bibinfo{title}{Attention is all you need}, in: \bibinfo{booktitle}{Advances
  in Neural Information Processing Systems 30: Annual Conference on Neural
  Information Processing Systems, {NIPS}}, \bibinfo{pages}{5998--6008},
  \bibinfo{year}{2017}.

\bibitem[{Viterbi(1967)}]{DBLP:journals/tit/Viterbi67}
\bibinfo{author}{A.~J. Viterbi}, \bibinfo{title}{Error bounds for convolutional
  codes and an asymptotically optimum decoding algorithm},
  \bibinfo{journal}{{IEEE} Trans. Information Theory}
  \bibinfo{volume}{13}~(\bibinfo{number}{2}) (\bibinfo{year}{1967})
  \bibinfo{pages}{260--269}.

\bibitem[{Artstein and Poesio(2008)}]{DBLP:journals/coling/ArtsteinP08}
\bibinfo{author}{R.~Artstein}, \bibinfo{author}{M.~Poesio},
  \bibinfo{title}{Inter-Coder Agreement for Computational Linguistics},
  \bibinfo{journal}{Comput. Linguistics}
  \bibinfo{volume}{34}~(\bibinfo{number}{4}) (\bibinfo{year}{2008})
  \bibinfo{pages}{555--596}.

\bibitem[{Liu et~al.(2014)Liu, Zhou, Wen, and
  Tang}]{DBLP:journals/ijmcmc/LiuZWT14}
\bibinfo{author}{Y.~Liu}, \bibinfo{author}{Y.~Zhou}, \bibinfo{author}{S.~Wen},
  \bibinfo{author}{C.~Tang}, \bibinfo{title}{A strategy on selecting
  performance metrics for classifier evaluation}, \bibinfo{journal}{{IJMCMC}}
  \bibinfo{volume}{6}~(\bibinfo{number}{4}) (\bibinfo{year}{2014})
  \bibinfo{pages}{20--35}.

\bibitem[{Zhou and Liu(2014)}]{zhou2014correlation}
\bibinfo{author}{Y.~Zhou}, \bibinfo{author}{Y.~Liu},
  \bibinfo{title}{Correlation analysis of performance metrics for classifier},
  in: \bibinfo{booktitle}{Decision Making and Soft Computing: Proceedings of
  the 11th International FLINS Conference}, \bibinfo{organization}{World
  Scientific}, \bibinfo{pages}{487--492}, \bibinfo{year}{2014}.

\bibitem[{Kingma and Ba(2015)}]{DBLP:journals/corr/KingmaB14}
\bibinfo{author}{D.~P. Kingma}, \bibinfo{author}{J.~Ba}, \bibinfo{title}{Adam:
  {A} method for stochastic optimization}, in: \bibinfo{booktitle}{3rd
  International Conference on Learning Representations, {ICLR}},
  \bibinfo{year}{2015}.

\bibitem[{Strubell et~al.(2017)Strubell, Verga, Belanger, and
  McCallum}]{DBLP:conf/emnlp/StrubellVBM17}
\bibinfo{author}{E.~Strubell}, \bibinfo{author}{P.~Verga},
  \bibinfo{author}{D.~Belanger}, \bibinfo{author}{A.~McCallum},
  \bibinfo{title}{Fast and accurate entity recognition with iterated dilated
  convolutions}, in: \bibinfo{booktitle}{Proceedings of the 2017 Conference on
  Empirical Methods in Natural Language Processing, {EMNLP}},
  \bibinfo{pages}{2670--2680}, \bibinfo{year}{2017}.

\bibitem[{He et~al.(2016)He, Zhang, Ren, and Sun}]{DBLP:conf/cvpr/HeZRS16}
\bibinfo{author}{K.~He}, \bibinfo{author}{X.~Zhang}, \bibinfo{author}{S.~Ren},
  \bibinfo{author}{J.~Sun}, \bibinfo{title}{Deep residual learning for image
  recognition}, in: \bibinfo{booktitle}{2016 {IEEE} Conference on Computer
  Vision and Pattern Recognition, {CVPR}}, \bibinfo{pages}{770--778},
  \bibinfo{year}{2016}.

\bibitem[{Wang et~al.(2015)Wang, Qian, Soong, He, and
  Zhao}]{DBLP:journals/corr/WangQSHZ15}
\bibinfo{author}{P.~Wang}, \bibinfo{author}{Y.~Qian}, \bibinfo{author}{F.~K.
  Soong}, \bibinfo{author}{L.~He}, \bibinfo{author}{H.~Zhao},
  \bibinfo{title}{Part-of-speech tagging with bidirectional long short-term
  memory recurrent neural network}, \bibinfo{journal}{arXiv preprint}
  \bibinfo{volume}{arXiv:1510.06168}.

\bibitem[{Huang et~al.(2015)Huang, Xu, and Yu}]{DBLP:journals/corr/HuangXY15}
\bibinfo{author}{Z.~Huang}, \bibinfo{author}{W.~Xu}, \bibinfo{author}{K.~Yu},
  \bibinfo{title}{Bidirectional LSTM-CRF models for sequence tagging},
  \bibinfo{journal}{arXiv preprint} \bibinfo{volume}{arXiv:1508.01991}.

\bibitem[{Wellner et~al.(2007)Wellner, Huyck, Mardis, Aberdeen, Morgan,
  Peshkin, Yeh, Hitzeman, and
  Hirschman}]{DBLP:journals/jamia/WellnerHMAMPYHH07}
\bibinfo{author}{B.~Wellner}, \bibinfo{author}{M.~Huyck},
  \bibinfo{author}{S.~A. Mardis}, \bibinfo{author}{J.~S. Aberdeen},
  \bibinfo{author}{A.~A. Morgan}, \bibinfo{author}{L.~Peshkin},
  \bibinfo{author}{A.~S. Yeh}, \bibinfo{author}{J.~Hitzeman},
  \bibinfo{author}{L.~Hirschman}, \bibinfo{title}{Research Paper: Rapidly
  Retargetable Approaches to De-identification in Medical Records},
  \bibinfo{journal}{{JAMIA}} \bibinfo{volume}{14}~(\bibinfo{number}{5})
  (\bibinfo{year}{2007}) \bibinfo{pages}{564--573}.

\bibitem[{Jiang et~al.(2017)Jiang, Zhao, He, Guan, and Jiang}]{Jiang_2017}
\bibinfo{author}{Z.~Jiang}, \bibinfo{author}{C.~Zhao}, \bibinfo{author}{B.~He},
  \bibinfo{author}{Y.~Guan}, \bibinfo{author}{J.~Jiang},
  \bibinfo{title}{De-identification of medical records using conditional random
  fields and long short-term memory networks}, \bibinfo{journal}{Journal of
  Biomedical Informatics} \bibinfo{volume}{75} (\bibinfo{year}{2017})
  \bibinfo{pages}{S43--S53}.

\end{thebibliography}

\end{CJK}
\end{document}